\documentclass[accepted]{uai2021} 

\usepackage[british]{babel}
\DeclareUnicodeCharacter{2212}{-}  
\usepackage{balance}
\usepackage{natbib} 
\usepackage{mathtools} 
\usepackage{booktabs} 
\usepackage{tikz} 

\title{Correlated Weights in Infinite Limits of Deep Convolutional Neural Networks}

\author[1]{\href{mailto:Adrià Garriga-Alonso <ag919@cam.ac.uk>?Subject=Your UAI 2021 paper}{Adrià~Garriga-Alonso}{}}
\author[2]{Mark~van~der~Wilk}
\affil[1]{%
    Department of Engineering\\
    University of Cambridge\\
    UK
}
\affil[2]{%
    Department of Computer Science\\
    Imperial College London\\
    UK
}

\usepackage{url}            
\usepackage{booktabs}       
\usepackage{amsfonts}       
\usepackage{nicefrac}       
\usepackage{microtype}      

\usepackage{xcolor}
\usepackage{amsfonts}

\usepackage{graphicx}
\usepackage{cancel}
\usepackage{import}  

\usepackage[colorinlistoftodos]{todonotes}

\usepackage{algorithm}
\usepackage[algo2e,ruled]{algorithm2e}

\usepackage{amsthm}
\usepackage{cleveref}  
\creflabelformat{equation}{#2#1#3}
\crefname{assumption}{assumption}{assumptions}
\usepackage{adria-math}  

\usepackage[single=true]{acro}
\DeclareAcronym{NN}{
  short = NN,
  long  = neural network,
}
\DeclareAcronym{DAG}{
  short = DAG,
  long  = directed acyclic graph,
}
\DeclareAcronym{RNN}{
  short = RNN,
  long  = recurrent neural network,
}
\DeclareAcronym{CNN}{
  short = CNN,
  long  = convolutional neural network,
}
\DeclareAcronym{LCN}{
  short = LCN,
  long  = locally-connected network,
}
\DeclareAcronym{FCNN}{
  short = FCNN,
  long  = fully connected neural network,
}
\DeclareAcronym{CLT}{
  short = CLT,
  long  = Central Limit Theorem,
}
\DeclareAcronym{RV}{
  short = RV,
  long  = random variable,
  long-plural-form = random variables,
}
\DeclareAcronym{iid}{
  short = i.i.d.,
  long  = independent and identically distributed,
}
\DeclareAcronym{GP}{
  short = GP,
  long  = Gaussian process,
  long-plural-form = Gaussian processes,
}

\DeclareAcronym{SE}{
  short = SE,
  long  = squared exponential,
}

\DeclareAcronym{ML}{
  short = ML,
  long  = machine learning,
}
\DeclareAcronym{BNN}{
  short = BNN,
  long  = Bayesian neural network,
}
\DeclareAcronym{ReLU}{
  short = ReLU,
  long  = rectified linear unit,
}
\DeclareAcronym{SGD}{
  short = SGD,
  long  = stochastic gradient descent,
}

\DeclareAcronym{CIFAR10}{
  short = CIFAR-10,
  long = CIFAR-10,  
}
\DeclareAcronym{MNIST}{
  short = MNIST,
  long = MNIST,  
}

\DeclareAcronym{GPU}{
  short = GPU,
  long = graphics processing unit,
}
\DeclareAcronym{GAP}{
  short = GAP,
  long = global average pooling,
}

\DeclareAcronym{MC}{
  short = MC,
  long = Monte Carlo,
}

\DeclareAcronym{MCMC}{
  short = MCMC,
  long = Markov chain Monte Carlo,
}

\DeclareAcronym{NTK}{
  short = NTK,
  long = neural tangent kernel,
}
\DeclareAcronym{DGP}{
  short = DGP,
  long = deep Gaussian process,
  long-plural-form = deep Gaussian processes,
}
\DeclareAcronym{DKL}{
  short = DKL,
  long = deep kernel learning,
}

\DeclareAcronym{MH}{
  short = MH,
  long = Metropolis-Hastings,
}

\DeclareAcronym{VI}{
  short = VI,
  long = variational inference,
}

\usepackage{xspace}

\newcommand{\Netsor}{\textsc{Netsor}\xspace}
\newcommand{\Gva}{\mathsf{G}}
\newcommand{\Hva}{\mathsf{H}}
\newcommand{\Ava}{\mathsf{A}}
\newcommand{\Ova}{\mathsf{O}}
\newcommand{\MatMul}{\texttt{MatMul}\xspace}
\newcommand{\LinComb}{\texttt{LinComb}\xspace}
\newcommand{\Nonlin}{\texttt{Nonlin}\xspace}
\newcommand{\SMatMul}[1]{{\MatMul: {#1}}}
\newcommand{\SLinComb}[1]{{\LinComb: {#1}}}
\newcommand{\SNonlin}[1]{{\Nonlin: {#1}}}

\newcommand{\SInput}{\Input}
\newcommand{\SOutput}{\Output}
\newcommand{\CommentC}{\tcc}

\newcommand{\Sigmax}{{\Sigma_{\text{x}}}}
\newcommand{\Sigmay}{{\Sigma_{\text{y}}}}
\newcommand{\Sigmaxy}{{\Sigma_{\text{xy}}}}

\newcommand{\layerAd}[2]{\mathbf{Z}^{(#1)}_{#2}(\vX')}
\newcommand{\layerAsd}[2]{Z^{(#1)}_{#2}(\vX')}
\newcommand{\layerAsm}[3]{Z^{(#1)}_{#2}\bra{#3}}
\newcommand{\layerAs}[2]{Z^{(#1)}_{#2}(\vX)}
\newcommand{\layerA}[2]{\mathbf{Z}^{(#1)}_{#2}(\mX)}
\newcommand{\layera}[2]{\mathbf{Z}^{(#1)}_{#2}(\mX)}
\newcommand{\layeram}[3]{\mathbf{Z}^{(#1)}_{#2}\bra{#3}}
\newcommand{\layerAm}[3]{\mathbf{Z}^{(#1)}_{#2}\bra{#3}}

\newcommand{\layerC}[1]{C^{(#1)}}

\newcommand{\layerNLAsd}[2]{A^{(#1)}_{#2}(\vX')}

\newcommand{\layerNLAs}[2]{A^{(#1)}_{#2}(\vX)}
\newcommand{\layerNLA}[2]{\vA^{(#1)}_{#2}(\vX)}

\newcommand{\layernlasm}[3]{\vA^{(#1)}_{#2}\bra{#3}}
\newcommand{\layernla}[2]{\vA^{(#1)}_{#2}(\mX)}

\newcommand{\layerU}[1]{\vU^{(#1)}}

\newcommand{\layerWs}[1]{W^{(#1)}}
\newcommand{\layerW}[1]{\mW^{(#1)}}

\newcommand{\layersizebase}{\vF}
\newcommand{\layersizebases}{F}
\newcommand{\layersize}[1]{{\layersizebase^{\bra{#1}}}}
\newcommand{\layerw}[1]{F_\text{w}^{(#1)}}
\newcommand{\layerh}[1]{F_\text{h}^{(#1)}}

\newcommand{\patchsizebase}{\vP}
\newcommand{\patchsizebases}{P}
\newcommand{\patchsize}[1]{{\patchsizebase^{\bra{#1}}}}

\newcommand{\patchw}[1]{P_\text{w}^{(#1)}}
\newcommand{\patchh}[1]{P_\text{h}^{(#1)}}

\newcommand{\patchf}[2]{{\tilde#1\bra{#2}}}
\newcommand{\convpatch}[1]{{\text{im}\bra{\tilde#1}}}

\newcommand{\priorWcovs}[1]{\Sigma^{(#1)}}
\newcommand{\priorWcov}[1]{\boldsymbol{\Sigma}^{(#1)}}

\newcommand{\chan}{i}
\newcommand{\prevchan}{j}   
\newcommand{\patch}{\vp}               
\newcommand{\patchs}{p}               
\newcommand{\nextpatch}{\vq}
\newcommand{\nextpatchs}{q}

\newcommand{\sigmaw}{{\sigma^{2}_{\text{w}}}}

\newcommand{\meanf}[1]{m\ssup{#1}}
\newcommand{\covf}[1]{K\ssup{#1}}
\newcommand{\nlinf}[1]{V\ssup{#1}}

\newcommand{\1}{\boldsymbol{1}}

\newcommand{\crefp}[1]{(\cref{#1})}
\newcommand{\eqparref}{\crefp}
\newcommand{\eqparreftwo}[2]{\crefp{#1,#2}}

\begin{document}
\maketitle

\begin{abstract}
Infinite width limits of deep neural networks often have tractable forms. They have been used to analyse the behaviour of finite networks, as well as being useful methods in their own right. When investigating infinitely wide convolutional neural networks (CNNs), it was observed that the correlations arising from spatial weight sharing disappear in the infinite limit.
This is undesirable, as spatial correlation is the main motivation behind CNNs. We show that the loss of this property is not a consequence of the infinite limit, but rather of choosing an independent weight prior. Correlating the weights maintains the correlations in the activations. 
Varying the amount of correlation interpolates between independent-weight limits and mean-pooling. Empirical evaluation of the infinitely wide network shows that optimal performance is achieved between the extremes, indicating that correlations can be useful.
\end{abstract}

\section{Introduction}
Analysing infinitely wide limits of neural networks has long been used to provide insight into the properties of neural networks.
\citet{neal1996bayesian} first noted such a relationship, through showing that infinitely wide Bayesian neural networks converge in distribution to Gaussian processes (GPs).
The success of GPs raised the question of whether such a comparatively simple model could replace a complex neural network. \citet{mackay1998introgp} noted that taking the infinite limit resulted in a fixed feature representation, a key desirable property of neural networks. Since this property is lost due to the infinite limit, MacKay inquired: ``have we thrown the baby out with the bath water?''

In this work, we follow the recent interest in infinitely wide convolutional neural networks \citep{garriga2018infiniteconv,novak2019infiniteconv}, to investigate another property that is lost when taking the infinite limit: correlation in the activations of patches in different parts of the image.
Given that convolutions were developed to introduce these correlations, and that they improve performance \citep{arora2019exact}, it seems undesirable that they are lost when more filters are added.
Currently, the only way of reintroducing spatial correlations is to change the model architecture by introducing mean-pooling \citep{novak2019infiniteconv}. This raises two questions:
\begin{enumerate}[label=\textbf{\arabic*)}]
\itemsep0em
\item Is the loss of patchwise correlations a necessary consequence of the infinite limit?
\item Is an architectural change the only way of reintroducing patchwise correlations?
\end{enumerate}

We show that the answer to both these questions is ``no''. Correlations between patches can also be
maintained in the limit without pooling by introducing correlations between the weights in the prior. The amount of correlation can be controlled,
which allows us to interpolate between the existing approaches of full
independence and mean-pooling. Our approach allows the discrete architectural choice of mean-pooling to be replaced with a more flexible continuous amount of correlation.

\begin{figure*}[t]
  \centering
{
  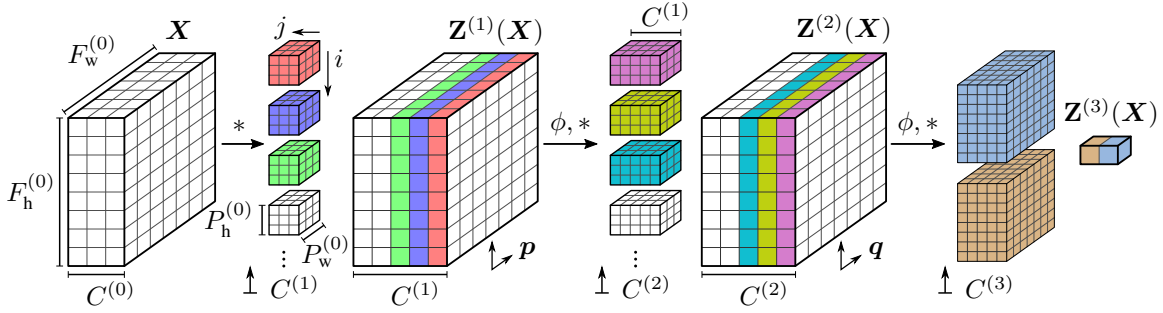}
  \caption{A deep convolutional neural network following our notation. Infinite limits are taken over the number of convolutional filters $\layerC{\ell}$ (vertical), which equals the number of channels in the following layer (horizontal). The network has $L=3$ layers and $D=2$ spatial dimensions. The output is not spatially extended $(\layersize{3}=\1)$ because $\patchsize{3} = \layersize{2}.$}
  \label{fig:fancy-cnn}
\end{figure*}

We empirically show that modest performance improvements can be obtained by replacing mean-pooling at the final layer with an intermediate amount of correlation. In addition, we show that in layers before the final one, the discrete architectural choice of mean-pooling can be replaced by an intermediate amount of correlation, without degrading performance. Avoiding discrete design decisions makes architecture search easier, by allowing continuous optimisation.
We speculate that these results from infinite networks could be useful for adapting priors or initialisations in finite networks, leading to better performance, or easier design.


Overall, our work illustrates that non-standard choices in the weight prior can significantly influence properties in the infinite limit, and that good choices can lead to improved performance. We hope that this work inspires investigation into correlated weights in finite neural networks, as well as more non-standard priors or initialisations.

\section{Spatial Correlations in Single~Hidden~Layer Networks}
\label{sec:single-layer}

To begin, we will analyse the infinite limit of a single hidden layer
convolutional neural network (CNN). This illustrates the choices that lead to the disappearance of spatial correlation in the activations. We extend \citet{garriga2018infiniteconv} and \citet{novak2019infiniteconv} by considering weight priors with correlations. By adjusting the correlation, we can interpolate between existing independent weight limits and mean-pooling, which previously had to be introduced as a discrete architectural choice. We also discuss how existing convolutional Gaussian processes \citep{vdw2017convgp,dutordoir2020} can be obtained from limits of correlated weight priors.


Consider a CNN with $L=2$ layers. \Cref{fig:fancy-cnn} provides a graphical representation of the notation.
The input $\mX$ is a real-valued tensor of shape ${\layerC{0} \times \layersize{0}}$, where $\layerC{0} \in \sN$ is the number of channels and $\layersize{0} \in \sN^{D}$ the spatial size of the input.  Superscripts denote the layer index.
For images, usually $\layerC{0}=3$ (one per colour), and the number of spatial input dimensions is $D=2$, so $\layersize{0} = (\layerh{0}, \layerw{0}).$
The convolution operation at layer $\ell\in\countto{L}$ divides its input into patches of size
$\patchsize{\ell} \preceq \layersize{\ell-1}$. For a given spatial location of the next activation
$\nextpatch\in\countto{\layersize{\ell}}$,\footnote{For some number
  $\patchsizebases\in \sN$, the expression $\countto{\patchsizebases}$ is the
  set $\cb{1,\dots,\patchsizebases}$. For a tuple $\patchsizebase\in\sN^{D}$,
  $\countto{\patchsizebase} = \countto{\patchsizebases_{1}}\times \dots \times \countto{\patchsizebases_{D}}$.}
 the patch function
$\patchf{\nextpatch}{\cdot}: \countto{\patchsize{\ell}} \to \countto{\layersize{\ell-1}}$
iterates over the elements of the patch \crefp{eq:little-patchf}. Weights are applied by taking an inner product with all patches, which we do $\layerC{\ell}$ times to give multiple channels in the next layer. By collecting all weights in the tensor $\layerW{\ell}\in\sR^{\layerC{\ell}\times \layerC{\ell-1}\times \patchsize{\ell}}$ the pre- and post-nonlinearity activations are respectively, for $\ell\in\countto{L},$
\begin{equation}
    \layerAs{\ell}{\chan,\nextpatch} = \sum_{\prevchan=1}^{\layerC{\ell-1}} \sum_{\patch=\1}^{\patchsize{\ell}} \layerWs{\ell}_{\chan,\prevchan,\patch}\, \layerNLAs{\ell-1}{\prevchan,\patchf{\nextpatch}{\patch}}\,,
    \label{eq:deep-recursion}
\end{equation}
\begin{equation}
  \layerNLAs{0}{\chan,\patch} = X_{\chan,\patch},\quad \layerNLAs{\ell}{\chan,\patch} = \phi\bra{\layerAs{\ell}{\chan,\patch}}\,.
  \label{eq:deep-activations}
\end{equation}
\Cref{eq:deep-recursion} is a channel-wise sum of $D$-dimensional convolutions, and $\phi$ denotes the elementwise nonlinearity.

For layer $\ell\in\countto{L}$, stride $s$, dilation $h$, the patch function is $\patchf{\nextpatch}{\patch} \eqdef \bra{\patchf{\nextpatchs_{1}}{\patchs_{1}}, \dots, \patchf{\nextpatchs_{D}}{\patchs_{D}}}$, where
\begin{equation}
  \patchf{\nextpatchs_{d}}{\patchs_{d}} = s\nextpatchs_{d} -  h\bra{\patchs_{d} - \ceil{\patchsizebases_{d}/2}}\,.
  \label{eq:little-patchf}
\end{equation}
Using \cref{eq:little-patchf}, it is possible to verify that \cref{eq:deep-recursion} is the usual deep learning convolution \crefp{app:patchf}.

In a single hidden layer CNN, these activations are followed by a fully-connected layer with weights $\layerW{2} \in \sR^{\layerC 1\times \patchsize{2}}$, where $\patchsize{2} = \layersize{1}$. Our final output is again given by a summation over the activations
\begin{align}
  f(\mX) &= \sum_{\prevchan=1}^{\layerC{1}} \sum_{\patch=1}^{\layersize{1}} \layerWs{2}_{\prevchan,\patch}\, \layerNLAs{1}{\prevchan,\patch}
         = \sum_{\patch=\1}^{\layersize{1}} \layerAs{f}{\patch}
        \label{eq:single-layer-f} \,,
\end{align}
where $\layerAs{f}{\patch}$ denotes the result before the summation over spatial locations $\patch$.

We analyse the distribution on function outputs $f(\vX)$ for some Gaussian prior
$p(\gW)$ on the weights of all layers $\gW$. In all the cases
we consider, we take the prior to be independent over layers and channels. Here
we extend earlier work by allowing spatial correlation in the final layer's
weights (we will consider all layers later) through the covariance tensor
$\priorWcov{\ell} \in \sR^{\layersize{\ell}\times\layersize{\ell}}$. This gives
the prior
\begin{align}
    p(\layerW1) &= \prod_{\chan=1}^{\layerC1}\prod_{\prevchan=1}^{\layerC0}\Normal{\layerW1_{\chan,\prevchan}\,;\, 0, \eye} \,, \\
     p(\layerW2) &= \prod_{\chan=1}^{\layerC1} \Normal{\layerW2_{\chan}\,;\, 0, \frac{1}{\layerC1}\priorWcov2} \,,
\end{align}
with independence between different layers' weights.
Here, a tensor-valued covariance $\priorWcov2$ expresses arbitrary covariance over the spatial dimensions of the tensor $\layerW{2}_{\chan}$: $\Cov\sqb{\layerWs{2}_{\chan,\patch},\layerWs{2}_{\chan,\patch'}} = \priorWcovs{2}_{\patch,\patch'}/\layerC{\ell-1}.$

Since $\layerW1,\layerW2$ are i.i.d. over channels $\chan\in\layerC{1}$, the random variables
$\sum_{\patch=\1}^{\layersize{1}}  \layerWs{2}_{\chan,\patch}\, \layerNLAs{1}{\chan,\patch}$ are identically distributed and independent for each $\chan \in \countto{\layerC{1}}$.
This allows us to apply the central limit theorem (CLT) to their sum $f(\mX)$, showing that $f(\mX)$ converges in distribution to a Gaussian process as $\layerC1\to\infty$ \citep{neal1996bayesian}.

The covariance between the final-layer activations for two inputs $\vX,\vX'$ becomes
\begin{align}
  &\Cov_{\vW}\left[\layerAs{f}{\patch},\layerAsd{f}{\patch'}\right] = \nonumber\\ &=\Expect{\vW}{\sum_{\prevchan=1}^{\layerC1}\sum_{\prevchan'=1}^{\layerC1} \layerWs{2}_{\prevchan,\patch}\layerNLAs{1}{\prevchan,\patch}\; \layerWs{2}_{\prevchan',\patch'}\layerNLAsd{1}{\prevchan',\patch'}}\,, \nonumber \\
  \intertext{use independences to split the expectations, and substitute the weight covariance,}
    &= \sum_{\prevchan=1}^{\layerC1}\sum_{\prevchan'=1}^{\layerC1} \Expect{\layerW1}{\layerNLAs{1}{\prevchan,\patch}\,\layerNLAsd{1}{\prevchan',\patch'}}\Expect{\layerW2}{\layerWs{2}_{\prevchan,\patch} \layerWs{2}_{\prevchan',\patch'}} \nonumber \\
  &= \sum_{\prevchan=1}^{\layerC1}\sum_{\prevchan'=1}^{\layerC1} \Expect{\layerW1}{\layerNLAs{1}{\prevchan,\patch}\,\layerNLAsd{1}{\prevchan',\patch'}}\;\delta_{\prevchan,\prevchan'}
    \frac{\priorWcovs{2}_{\patch,\patch'}}{\layerC{1}}\,,\nonumber \\
  \intertext{eliminate one of the sums over $\prevchan$ using $\delta_{\prevchan,\prevchan'}$, and rearrange}
  &= \Expect{\layerW1}{
\frac{1}{\layerC1}\sum_{\prevchan=1}^{\layerC1} \layerNLAs{1}{\prevchan,\patch}\,\layerNLAsd{1}{\prevchan,\patch'}}
    \priorWcovs{2}_{\patch,\patch'} \nonumber \\
  &=  \nlinf{1}_{\patch,\patch'}\bra{\mX,\mX'}  \;\priorWcovs{2}_{\patch,\patch'} \,.
\end{align}
The limit of the sum of the final expectation over $\layerW1$ can be found in closed form for many activations (see \cref{sec:exp-nonlin}) and is denoted $\nlinf{1}_{\patch,\patch'}\bra{\mX,\mX'}.$
Note in \cref{eq:deep-recursion} that the activations for some location $\patch\in\countto{\layersize{1}}$ only depend on the input patch at $\patch$, that is, on the elements of $\mX$ that are in the image $\convpatch{\patch}$ of the patch function $\patchf{\patch}{\cdot}$. Thus, the kernel acts locally on patches:
$\nlinf{1}_{\patch,\patch'}\bra{\mX,\mX'} = k\ssup{1}\bra{\mX_{:,\,\convpatch{\patch}},\mX'_{:,\,\convpatch{\patch'}}}$.

We find the final kernel for the GP by taking the covariance between function values $f(\vX)$ and $f(\vX')$ and performing the final sum in \cref{eq:single-layer-f}:
\begin{align}
    K(\vX, \vX') &= \Cov\left[f(\vX), f(\vX')\right] \nonumber\\
    &= \sum_{\patch,\patch'} k\ssup{1}\bra{\mX_{:,\,\convpatch{\patch}},\mX'_{:,\,\convpatch{\patch'}}}\, \priorWcovs{2}_{\patch,\patch'} \label{eq:cov-kernel} \,.
\end{align}
We can now see how different choices for $\priorWcov2$ give different forms of spatial correlation.

\paragraph{Independence.} \Citet{garriga2018infiniteconv} and \citet{novak2019infiniteconv} consider $\priorWcovs{2}_{\patch,\patch'} = \delta_{\patch,\patch'}\sigmaw$,
i.e.~the case where all weights are independent. The resulting kernel simply sums components over patches, which implies an \emph{additive model} \citep{stone1985}, where a \emph{different} function is applied to each patch, after which they are all summed together: $f(\vX) = \sum_\patch f_\patch(\vX_{:,\convpatch{\patch}})$.
This structure has commonly been applied to improve GP performance in high-dimensional settings \citep[e.g.][]{duvenaud2011additive,durrande2012additive}. \citet{novak2019infiniteconv} point out that the same kernel can be obtained by taking an infinite limit of a \emph{locally connected network} (LCN) \citep{lecun1989generalization} where connectivity is the same as in a CNN, but without weight sharing, indicating that a key desirable feature of CNNs is lost.

\paragraph{Mean-pooling.} By taking $\priorWcovs{2}_{\patch,\patch'} = 1/\abs{\layersize{2}}^{2}$ we make the weights fully correlated over all locations\footnote{For a size $\layersizebase \in \sN^{D}$, its number of elements is $\abs{\layersizebase} \eqdef \prod_{d=1}^{D}\layersizebases_{d}.$}, leading to identical weights for all $\patch$, i.e.~$\layerWs{2}_{\chan,\patch} = \layerWs{2}_{\chan}$. This is equivalent to taking the mean response over all spatial locations (see \cref{eq:single-layer-f}), or global average pooling. As \citet{novak2019infiniteconv} discuss, this reintroduces the spatial correlation that is the intended result of weight sharing. The ``translation invariant'' convolutional GP
of \citet{vdw2017convgp} can be obtained by this single-layer limit using Gaussian activation functions \citep{vdw2019thesis}. Since this mean-pooling was shown to be too restrictive in this single-layer case, \Citet{vdw2017convgp} considered pooling with constant weights $\alpha_\patch$ (i.e.~without a prior on them). In this framework, this is equivalent to placing a rank 1 prior on the final-layer weights by taking $\priorWcovs{2}_{\patch\patch'} = \alpha_\patch \alpha_{\patch'}$. This maintains the spatial correlations, but requires the $\alpha_\patch$ parameters to be learned by maximum \textit{marginal} likelihood (ML-II, empirical Bayes).


\paragraph{Spatially correlated weights.} In the pooling examples above, the spatial covariance of weights is taken to be a rank-1 matrix. We can add more flexibility to the model by varying the strength of correlation between weights based on their distance in the image. We consider an exponential decay depending on the distance between two patches: $\priorWcovs{2}_{\patch\patch'} = \exp\bra{-d\bra{\patch, \patch'}/l}$. We recover full independence by taking $l\to 0$, and mean-pooling with $l\to\infty$. Intermediate values of $l$ allow the rigid assumption of complete weight sharing to be relaxed, while still retaining spatial correlations between similar patches. This construction gives the same kernel as investigated by \citet{mairal2014ckn} and \citet{dutordoir2020}, who named this property ``translation insensitivity'', as opposed to the stricter invariance that mean-pooling gives. The additional flexibility improved performance without needing to add many parameters that are learned in a non-Bayesian fashion.

Our construction shows that spatial correlation can be retained in infinite limits without needing to resort to architectural changes. A simple change to the prior on the weights is all that is needed. This property is retained in wide limits of deep networks, which we investigate next.



\section{Spatial Correlations in Deep Networks}
\label{sec:deep-correlations}
Here, we provide an informal extension of the previous section's results to deep networks.
In deep networks, correlated weights also retain spatial correlation in the activations.
\Cref{app:netsor} provides a formal justification for this section, using the framework by \citet{yang2019wide}.

The procedure for computing the kernel has a recursive form similar to existing analyses \citep{garriga2018infiniteconv,novak2019infiniteconv}. Negligible additional computation is needed to consider arbitrary correlations, compared to only considering mean-pooling \citep{novak2019infiniteconv,arora2019exact}. The main bottleneck is the need for computing covariances for all pairs of patches in the image, as in \cref{eq:cov-kernel}. For a $D$-dimensional convolutional layer, the corresponding kernel computation is a convolution of the activations' second moment with the
$2D$-dimensional covariance tensor of the weights.

The setup for the case of a deep neural network follows that of \cref{sec:single-layer}, but with the number of layers $L>2$. The outputs of the network are simply the pre-nonlinearity activations of the $L$th layer, $\layerAs{L}{\chan,\patch}$. If we need several outputs, for example in $K$-class classification, we may set $\layerC{L}=K.$
If the output of the network should not be spatially extended, we set the spatial size to $\layersize{L}=\1$. This can be achieved by making the weights $\layerW{L}$ (and their corresponding convolutional patch) have the same size as $\layersize{L-1}$ (see \cref{fig:fancy-cnn}).

As pointed out by \citet{matthews2018dnnlimit}, a straightforward application of the central limit theorem is not possible for deep networks. Fortunately, \citet{yang2019wide} developed a general framework for expressing neural network architectures and finding their corresponding Gaussian process infinite limits. The resulting kernel is given by the recursion that can be derived from a more informal argument which takes the infinite width limit in a sequential layer-by-layer fashion, as was used in \citet{garriga2018infiniteconv}. We follow this informal derivation, as this more naturally illustrates the procedure for computing the kernel. A formal justification can be found in \cref{app:netsor}.

\subsection{Recursive Computation of the Kernel}
In our weight prior, we correlate weights \emph{within} a convolutional filter. The weights remain independent over layers and channels. For each $\ell \in \countto{L}$,
\begin{equation}
  p\bra{\layerW{\ell}} = \prod_{\chan=1}^{\layerC{\ell}}\prod_{\prevchan=1}^{\layerC{\ell-1}}
  \Normal{\layerW{\ell}_{\chan,\prevchan}; 0, \frac{1}{\layerC{\ell-1}} \priorWcov{\ell}}.
    \label{eq:correlated-weight-prior}
\end{equation}
As in \cref{sec:single-layer}, $\layerW{\ell} \in \sR^{\layerC{\ell}\times\layerC{\ell-1}\times\layersize{\ell}}$, and the covariance tensor $\priorWcov{\ell} \in \sR^{\layersize{\ell}\times\layersize{\ell}}$ is positive semi-definite.
Our derivation is general for any weight covariance, so layers with correlated weights can be interspersed with the usual layers.

A Gaussian process is determined by the mean and covariance of function values for pairs of inputs $\vX,\vX'$. The mean is zero. Using the recursion in \cref{eq:deep-recursion}, we can find the covariance between any two pre-nonlinearity activations from a pair of inputs $\vX,\vX'$ and the covariance of the previous layer. For $\chan,\chan'\in\countto{\layerC{\ell}}$ and $\nextpatch,\nextpatch'\in\countto{\layersize{\ell}}$,
\begin{align}
  &\Cov_{\gW}\left[\layerAs{\ell}{\chan,\nextpatch},\layerAsd{\ell}{\chan',\nextpatch'}\right] = \nonumber\\
  &=\sum_{\prevchan,\prevchan',\;\patch,\patch'} \Expect{\gW}{
    \layerWs{\ell}_{\chan,\prevchan,\patch}\layerNLAs{\ell-1}{\prevchan,\patchf{\nextpatch}{\patch}}\; \layerWs{\ell}_{\chan',\prevchan',\patch'}\layerNLAsd{\ell-1}{\prevchan',\patchf{\nextpatch'}{\patch'}}}\,, \nonumber \\
  \intertext{substituting the expression for the weight covariance,}
  &= \delta_{\chan,\chan'} \sum_{\patch,\patch'}
    \priorWcovs{\ell}_{\patch,\patch'}
    \E_{\gW} \bigg[\nonumber\\
  &\hspace{4em}
 \frac{1}{\layerC{\ell-1}}\sum_{\prevchan=1}^{\layerC{\ell-1}}
    \layerNLAs{\ell-1}{\prevchan,\patchf{\nextpatch}{\patch}}\,
    \layerNLAsd{\ell-1}{\prevchan,\patchf{\nextpatch'}{\patch'}}
    \bigg] \nonumber \\
  &= \delta_{\chan,\chan'} \covf{\ell}_{\nextpatch,\nextpatch'}\bra{\mX,\mX'}\,.\label{eq:kernel-recursion}
\end{align}
We can see that the covariance of activations in different channels
$(\chan\ne\chan')$ is zero. Otherwise, to calculate
$\covf{\ell}\bra{\mX,\mX'}$, we need to calculate the
expectation over $\gW$, which we term $\nlinf{\ell-1}\bra{\mX,\mX'}.$ The resulting kernel expression is
\begin{equation}
  \covf{\ell}_{\nextpatch,\nextpatch'}\bra{\mX,\mX'} = \sum_{\patch=\1}^{{\patchsize{\ell}}}
  \sum_{\patch'=\1}^{{\patchsize{\ell}}} \priorWcovs{\ell}_{\patch,\patch'}
  \nlinf{\ell-1}_{\patchf{\nextpatch}{\patch},\patchf{\nextpatch'}{\patch'}}\bra{\mX,\mX'}\,,
  \label{eq:cnn-kernel-recursive}
\end{equation}
which, because the concatenation of patch functions is a patch function \crefp{rem:concat-patchf}, is equivalent to a $2D$-dimensional convolution. This kernel does not correspond to a locally-connected network, because it uses off-diagonal elements of the previous layer's kernel.

\subsection{Expectation of the Nonlinearities}
\label{sec:exp-nonlin}
For $\ell=0$, the activations in the previous layer are the image inputs, i.e.~$\layerNLA{0}{} = \vX$ (\cref{eq:deep-activations}), making $\nlinf{0}_{\nextpatch,\nextpatch'}\bra{\mX,\mX'}$ an inner product between image patches.

For $\ell\ge 1$, the expression inside the expectation in \cref{eq:kernel-recursion} is a random variable, an average over $\prevchan\in\countto{\layerC{\ell-1}}$. From \cref{eq:deep-recursion} we see that all its terms have the same expectation, i.e.
\begin{multline}
\nlinf{\ell}_{\patch,\patch'}\bra{\mX,\mX'}
= \Expect{\gW}{\frac{1}{\layerC{\ell}} \sum_{\prevchan=1}^{\layerC{\ell}} \layerNLAs{\ell}{\prevchan,\patch} \layerNLAsd{\ell}{\prevchan,\patch'}} \\
= \Expect{ \layerA{\ell}{}, \layerAd{\ell}{} }{\phi\bra{\layerAs{\ell}{1,\patch}} \phi\bra{\layerAsd{\ell}{1,\patch'}}}\,.
\label{eq:avg-to-gaussian}
\end{multline}
For the purposes of \cref{eq:avg-to-gaussian}, in the infinite width limit, the
pre-nonlinearities $\layerA{\ell}{},\layerAd{\ell}{}$ converge in distribution to a joint Gaussian
\crefp{thm:netsor-master}. Accordingly, the value of the expectation above depends only on the entries of their $2\times 2$ covariance matrix and the form of $\phi$. Here we represent this dependence through the function $F_{\phi}(\Sigmax, \Sigmay, \Sigmaxy)$,
\begin{multline}
\nlinf{\ell}_{\patch,\patch'}\bra{\mX,\mX'}
=F_{\phi}\big(\\
\covf{\ell}_{\patch,\patch'}\bra{\mX, \mX},\,\covf{\ell}_{\patch,\patch'}\bra{\mX', \mX'},
\covf{\ell}_{\patch,\patch'}\bra{\mX, \mX'}\big)\,.
\label{eq:post-nonlin}
\end{multline}

Combining \cref{eq:cnn-kernel-recursive},~\ref{eq:post-nonlin} and the input inner product
provides us with a recursive procedure to compute the covariances all the way up to the final layer.

For the balanced ReLU nonlinearity ($\phi(x) = \sqrt{2}\max(0, x)$), which we
use in all the experiments in this paper, we can use the expression by \citet{cho2009mkm}:
\begin{multline}
  F_{\phi}\bra{\Sigmax, \Sigmay, \Sigmaxy} = \frac{1}{\pi} \sqrt{\Sigmax\Sigmay - \Sigmaxy^{2}} \\+ \bra{1 - \frac{1}{\pi}\cos^{-1}\bra{\frac{\Sigmaxy^{2}}{\Sigmax\Sigmay}}} \Sigmaxy.
\end{multline}

This expression implies that
$\nlinf{\ell}_{\patch,\patch}\bra{\mX,\mX} = \covf{\ell}_{\patch,\patch}\bra{\mX,\mX}$,
for all $\mX$ and $\patch$ \citep{lee2018dnnlimit,matthews2018dnnlimit}.

\subsection{Computational complexity, diagonal propagation}
To handle the covariance tensor for $\covf{\ell}\bra{\mX,\mX'}$, we need to compute and represent $\absshort{\layersize{\ell}}^{2}$ entries. This can be considerably more expensive than the forward pass of the corresponding CNN, where the activations have size $\absshort{\layersize{\ell}}$.
In special cases, the computation or memory costs can be reduced, compared to the $2D$-dimensional convolution in \cref{eq:cnn-kernel-recursive}, which is a generalisation of previous algorithms. These cases do not include layers with mean-pooling, for which our algorithm is equally expensive to previous ones \citep{arora2019exact}.

If weights are independent, only the diagonal of $\priorWcov{\ell}$ has nonzero entries, so $\priorWcovs{\ell}_{\patch,\patch'} = \delta_{\patch,\patch'}\priorWcovs{\ell}_{\patch,\patch'}$. One of the sums in the \cref{eq:cnn-kernel-recursive} can then be removed,
\begin{equation}
\covf{\ell}_{\nextpatch,\nextpatch'}\bra{\mX, \mX'} =
\sum_{\patch=\1}^{\patchsize{\ell-1}}
\priorWcovs{\ell}_{\patch,\patch} \,
\nlinf{\ell-1}_{\patchf{\nextpatch}{\patch},\patchf{\nextpatch'}{\patch}}\bra{\mX, \mX'}.
\label{eq:cnn-kernel-independent-weights}
\end{equation}
The patch functions that access $\nlinf{\ell-1}\bra{\mX,\mX'}$ are still
different ($\patchf{\nextpatch}{\cdot}$ and $\patchf{\nextpatch'}{\cdot}$), but
their argument $\patch$ is the same.

Patch functions \crefp{def:patchf} subtract their argument multiplied by the dilation. Consequently, the difference  of two patch functions with the same argument is constant: $\patchf{\nextpatch}{\patch} - \patchf{\nextpatch'}{\patch} = s\cdot\bra{\nextpatch - \nextpatch'}$. This means that the terms of the sum are on the same diagonal. Thus, to calculate the covariance for a given location pair $\nextpatch,\nextpatch'$, we need to do a single sum \emph{over a diagonal} of the second moment tensors $\nlinf{\ell-1}\bra{\mX, \mX'}$ and $\priorWcov{\ell}.$

This results in exact same algorithm as \citet{arora2019exact}, which convolves over the diagonals, for layers with independent weights. Its memory cost is still $O(\absshort{\layersize{\ell}}^{2})$, but the computational cost is reduced to $O(\absshort{\layersize{\ell}}^{2} \absshort{\patchsize{\ell}})$, compared to $O(\absshort{\layersize{\ell}}^{2} \absshort{\patchsize{\ell}}^{2})$ for non-diagonal covariance.

\paragraph{Diagonal propagation with independent weights.}
Exactly \emph{which} diagonal of $\nlinf{\ell-1}\bra{\mX, \mX'}$ do we need to sum over? Clearly, it is the one indexed by $s\cdot\bra{\nextpatch - \nextpatch'}$, i.e.~the one that contains the position $\bra{s\nextpatch, s\nextpatch'}$. Thus, the number of diagonals of $\nlinf{\ell-1}\bra{\mX, \mX'}$ that we will need to access is exactly the number of possible values that $\nextpatch-\nextpatch'$ can take. That number is determined by the size $\layersize{\ell}$ of layer $\ell$, but is completely unrelated to the size $\layersize{\ell-1}$ of layer $\ell-1$.

Fix some layer $\ell\in\countto{L}$.
We can iterate this argument from layer $\ell$ to layer $1$ to show that, for all $m \le \ell$, the number of diagonals of $\covf{m}\bra{\mX,\mX'}$ that one needs to calculate depends only on $\layersize{\ell}$. This can yield significant computational savings when the stride is $s\ge 2$ for one or more layers.

\paragraph{Last layer not spatially extended.} When the last layer is not spatially extended, its size is $\layersize{L}=\1$, so it only has one diagonal. If all the weights of the CNN are independent, this implies that we only need to calculate one diagonal of the covariance for every layer. That is:
\begin{equation}
\covf{\ell}_{\nextpatch,\nextpatch}\bra{\mX, \mX'} =
\sum_{\patch=\1}^{\patchsize{\ell-1}}
\priorWcovs{\ell}_{\patch,\patch} \,
\nlinf{\ell-1}_{\patchf{\nextpatch}{\patch},\patchf{\nextpatch}{\patch}}\bra{\mX, \mX'}.
\label{eq:cnn-kernel-independent-diagonal}
\end{equation}
With this simplification, the convolutions required to calculate the kernel are
$D$-dimensional, bringing the memory cost to $O(\absshort{\layersize{\ell}})$ and computational cost to
$O(\absshort{\layersize{\ell}}\absshort{\patchsize{\ell}})$, same as the finite CNN \citep{garriga2018infiniteconv}. The resulting kernel is equivalent to that of a
locally connected network.

\subsection{Implementation}
We extend the \texttt{neural-tangents} \citep{neuraltangents2020} library with a convolution layer and a fully connected layer, that admit a 4-dimensional covariance tensor for the weights. This allows interoperation with existing layers.

Since 4d convolutions are uncommon in deep learning, our implementation uses a sum over $\patchh{\ell}$ 3-d convolutions, where $\patchh{\ell}=3$ is the spatial height of the convolutional filter. While this enables GPU acceleration, computing the kernel is a costly operation. Reproducing our results takes around 10 days using an nVidia RTX 2070 GPU. Access to computational resources limited our experiments to subsets of data on CIFAR-10.

\section{Experiments}
By considering different amounts of correlation, we can interpolate between existing architectures that use independent weights or full mean-pooling. We consider two possible benefits of using this larger, continuously parameterised space of models:
\begin{enumerate}[label=\textbf{\arabic*)}]
    \item Decreased reliance on discrete architectural choices like mean-pooling.
    \item Improved performance by finding a better model in the expanded search space.
\end{enumerate}

Discrete choices pose a challenge for architecture search, as a separate network needs to be trained to evaluate the effect of each choice, which is computationally expensive. Continuous choices are preferable, as gradients can often be used to adjust many choices simultaneously. We investigate whether the discrete choice of mean-pooling can instead be replaced by a suitable selection of the continuous correlation parameter in a larger convolutional filter. While searching in this larger space of kernels, we also hope to observe improved performance. We investigate these two questions by performing parameter search in the next two sections.



\subsection{Experimental setup}
We evaluate various models on class-balanced subsets of CIFAR-10 of size $2^i \cdot 10$, following \citet{arora2020small}. As is
standard practice in the wide network literature, we reframe classification as
regression to one-hot targets $\vY$. 
We subtract $C=0.1$ from $\vY$ to make its mean zero, but we observed that this affects the results very little. The prediction is the class $k$ with highest mean of the posterior Gaussian process
\begin{align}
\text{label}(x_*) &= \text{argmax}_k \,f_k(x_*) \nonumber \\
&= \text{argmax}_k \,\vK_{x_* \vX}\bra{\sigma^2\eye + \vK_{\vX\vX}}^{-1}\vY_{:,k}\,,
\label{eq:mean-gp}
\end{align}
where $\sigma^2$ is a hyperparameter, the variance of the observation noise of the GP regression. We perform cross-validation to find a setting for $\sigma^2$. We use the eigendecomposition of $\vK_{xx}$ to avoid the need to recompute the inverse for each value of $\sigma^2$.

In the next two experiments we investigate the cross-validation performance on subsets of CIFAR-10 for a sweep of correlation parameters on two different neural network architectures.
We consider two architectures used in the neural network kernel literature, the CNN-GP \citep{novak2019infiniteconv,arora2019exact} with 14 layers, and the Myrtle network \citep{shankar2020without} with 10 layers. The CNNGP-14 architecture $((\texttt{conv}, \texttt{relu})\times 14, \texttt{pool})$ has a $32 \times 32$-sized layer at the end, which is usually transformed into the $1 \times 1$ output using global average pooling. The Myrtle10 architecture $(((\texttt{conv},\texttt{relu})\times 2, \texttt{pool}_{2\times 2}) \times 3, \texttt{pool})$ has a $8\times 8$ pooling layer at the end.



\begin{figure*}
\centering
 \scalebox{1.0}{\input{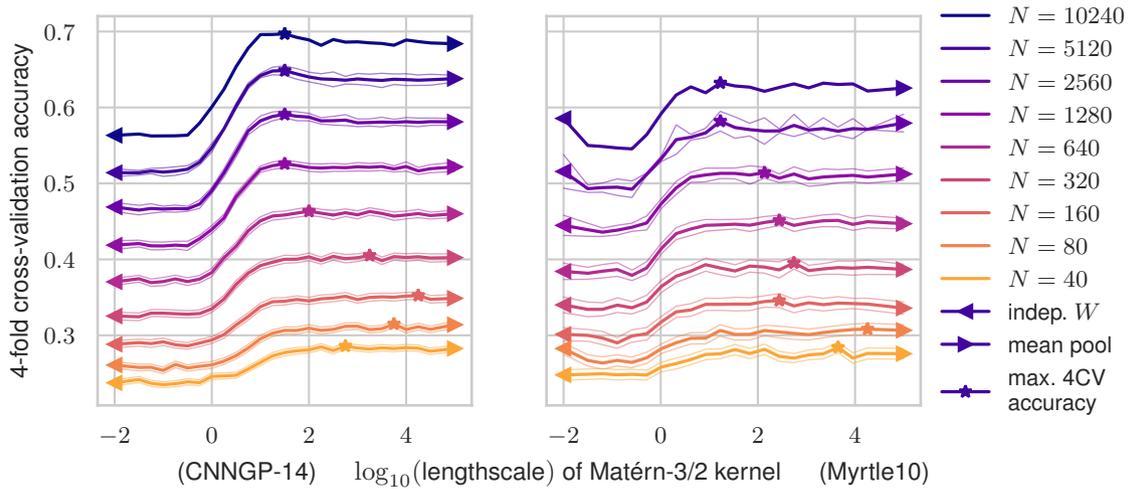}}
  \caption{Cross-validation accuracy of the CNNGP-14 and Myrtle10 networks on subsets of CIFAR10, with varying lengthscale of the Matérn-$3/2$ kernel that determines the weight correlation in the last layer. With larger data set sizes $N$, the improvement is larger, and the optimal lengthscale $\lambda$ converges to a similar value $(\lambda \approx 17)$. For all data sets except the largest, the values are averaged over several runs, and the thin lines represent the $\pm 2\sigma_n$, the estimated standard deviation of the mean. We can improve the performance of the classifier by choosing an intermediate $\lambda$. \label{fig:last-layer}}
\end{figure*}

\begin{figure*}[htpb]
  \centering
  \scalebox{1.0}{\input{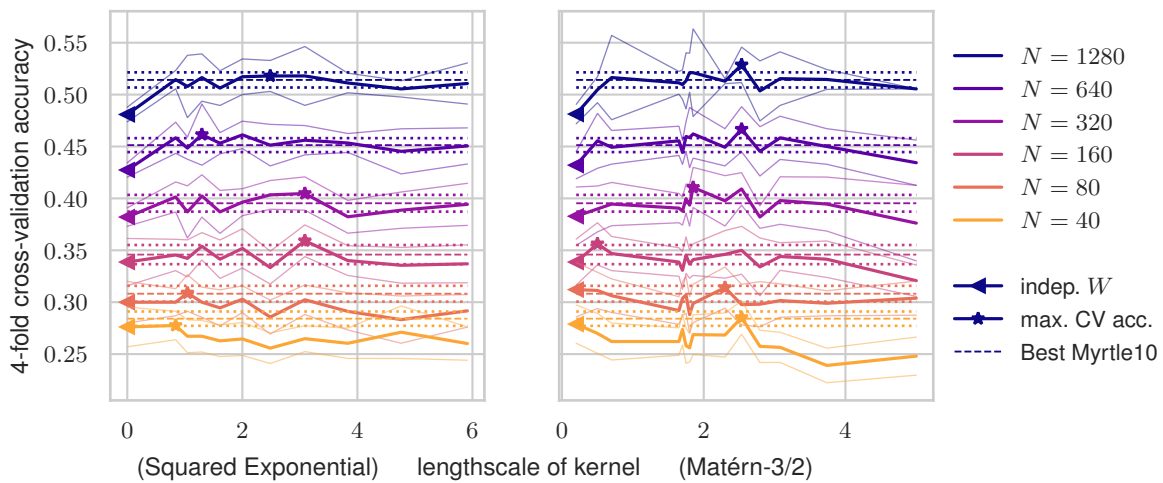}}.
  \caption{Correlated weights in intermediate layers. We replace pooling layers in the Myrtle10 architecture with larger convolutional filters with correlated weights. The lengthscale, and thus the amount of correlation, is varied along the x-axis. By adding correlations to a convolutional layer, we can recover (but not, in this case, exceed) the performance of the hand-selected architecture with mean-pooling. \label{fig:all-layers}}
\end{figure*}

\subsection{Correlated weights in the last layer}
We begin by investigating the addition of correlations in the weights of the final layer, since this is sufficient to prevent the disappearance of spatially correlated activations.  Following \citet{dutordoir2020}, the covariance $\vSigma_{pp'}$ of the weights is given by the Matérn-3/2 kernel with lengthscale $\lambda$:
\begin{equation}
    \priorWcovs{L}_{\patch,\patch'} = \bra{1 + \frac{\sqrt{3}\bracket{\|}{\|}{\patch - \patch'}_2}{\lambda}} \exp\bra{-\frac{\sqrt{3}\bracket{\|}{\|}{\patch - \patch'}_2}{\lambda}}.
\end{equation}
where we see the patch locations $\patch,\patch'$ as vectors. The ``extremes''
of independent weights and mean pooling are represented by
$\priorWcovs{L}_{\patch,\patch'} = \delta_{\patch,\patch'}$ and
$\priorWcovs{L}_{\patch,\patch'} = 1$, respectively. \footnote{Strictly
  speaking, $\priorWcov{L} = {\boldsymbol 1\boldsymbol 1}^\tp$ corresponds to
  sum-pooling, but the missing constant $\absshort{\layersizebase}^{-2}$ does
  not affect the maximum in \cref{eq:mean-gp}.}

\Cref{fig:last-layer}, shows how the 4-fold cross-validation accuracy on the training sets varies with the lengthscale $\lambda$ of the Matérn-3/2 kernel, which controls the ``amount'' of spatial correlation in the weights of the last layer. For each data point in each line, we split the data set into 4 folds, and we calculate the test accuracy on 1 fold using the other 3 as training set, for each value of $\sigma$ that we try. We take the maximum accuracy over $\sigma$.

We investigate how the effect above varies with data set size. The results in \cref{fig:last-layer} show that particularly for the CNNGP-14 architecture, correlated weights in the final layer lead to a modest but consistent improvement in performance, with the effect becoming larger with increasing dataset size. We can also see the optimal lengthscale $\lambda$  converging to a similar value for both architectures, of about $\lambda \approx 17$, which is evidence that the improvement holds for larger data sets. The optimal lengthscale is the same for both networks, so we speculate it may be a property of the {CIFAR10} data set.

\paragraph{Data partitioning.} The largest data set size in each part of the plot was run only once because of computational constraints. We transform one data set of size $N$ into two data sets of size $N/2$ by taking block diagonals of the stored kernel matrix, so we have more runs for the smallest sizes. This is an unbiased \acl{MC} estimate of the true accuracy under the data distribution, since the individual data points are uniformly distributed (but not independent, since they are sampled without replacement). It also has less variance than independent data sets, because the data sets taken are anti-correlated; they have no points in common. Accordingly, the error bars in \cref{fig:last-layer,fig:all-layers} are an estimate of the standard error: the square root of an upwards-biased estimator of the variance of the mean.

\paragraph{Implementation.} We use the \texttt{neural-tangents} \citep{neuraltangents2020} library to calculate the spatial kernel at the previous-to-last layer, $\covf{L-1}\bra{\mX,\mX'}$, once.
Since only the lengthscale of the last layer changes, we can cheaply obtain the final layer kernel matrix $\vK\ssup{L}_{\vX\vX'}$ for all lengthscales.

\subsection{Correlated weights in intermediate layers}
We take the same approach to the experiment in \cref{fig:all-layers}. To investigate whether correlated weights can replace mean-pooling, we replace the $2\times 2$ intermediate mean-pooling layer, together with the next $3\times 3$ convolution layer, in the Myrtle10 architecture with correlated weights. We change them to a $6 \times 6$ weight-correlated convolution. We vary the lengthscale for the covariance of all the newly correlated layers, setting them to the same value.

We observe that for independent weights (lengthscale is 0) the performance of the network is significantly below the optimum. Correlating the weights improves performance, although after adding small amounts of correlation, performance stays roughly constant. This indicates that for intermediate layers mean-pooling is not a sub-optimal choice, as it is for the last layer. However, the amount of correlation is a continuous parameter, which could lead to avoiding this discrete choice in model architecture.

\paragraph{Implementation.}
In this experiment, the lengthscales vary across the whole network, so we need to calculate $\covf{L-1}\bra{\mX,\mX'}$ every time. For a given data set size, this makes each point in \cref{fig:all-layers} considerably more expensive. For each data point, we optimise over the lengthscale of the last layer like in \cref{fig:last-layer}, picking the one with highest cross-validation accuracy.

\section{Related work}
Infinitely wide limits of neural networks are currently an important tool for creating approximations and analyses. Here we provide a background on the different infinite limits that have been developed, together with a brief overview of where they have been applied.

Interest in infinite limits first started with research into properties of Bayesian priors on the weights of neural networks. \citet{neal1996bayesian} noted that prior function draws from a single hidden layer neural network with appropriate Gaussian priors on the weights tended to a Gaussian process as the width grew to infinity. The simplicity of performing Bayesian inference in Gaussian process models led to their widespread adoption soon after \citep{williams1996gpr,gpml}. Over the years, the wide limits of networks with different weight priors and activation functions have been analysed, leading to various \emph{kernels} which specify the properties of the limiting Gaussian processes \citep{williams1997inf,cho2009mkm}.

With the increasing prominence of deep learning, recursive kernels were introduced in an attempt to obtain similar properties. \citet{cho2009mkm,mairal2014ckn} investigated such methods for fully-connected and convolutional architectures respectively. Despite similarities between recursive kernels and neural networks, the derivation did not provide clear relationships, or any equivalence in a limit. \citet{hazan2015} took initial steps to showing the wide limit equivalence of a neural network beyond the single layer case. Recently, \citet{matthews2018dnnlimit,lee2018dnnlimit} simultaneously provided general results for the convergence of the prior of deep fully-connected networks to a GP.\footnote{The derivation of the limiting kernel differs between the two papers, with the results being consistent. \citet{matthews2018dnnlimit} carefully take limits of realisable networks, while \citet{lee2018dnnlimit} take the infinite limit of each layer sequentially.
} 
A different class of limiting kernels, the Neural Tangent Kernel (NTK), originated from analysis of the function implied by a neural network during optimisation \citep{jacot2018ntk}, rather than the prior implied by the weight initialisation. Just like the Bayesian prior limit, this kernel sheds light on certain properties of neural networks, as well as providing a method with predictive capabilities of its own. 
The two approaches end up with subtly different kernels, which both can be computed as a recursive kernel. Both such infinite limits have recently been used for predicting and analysing training properties of finite neural networks \citep{poole2016,schoenholz2017deepinformationpropagation,hayou2019activation}, as well as for (Bayesian) training of infinitely wide networks.

With the general tools in place, \citet{garriga2018infiniteconv,novak2019infiniteconv} derived limits of the prior of convolutional neural networks with infinite filters. 
These two papers directly motivated this work by noting that spatial correlations disappeared in the infinite limit. Spatial mean pooling at the last layer was suggested as one way to recover correlations, with \citet{novak2019infiniteconv} providing initial evidence of its importance. Due to computational constraints, they were limited to using a Monte Carlo approximation to the limiting kernel, while \citet{arora2019exact} performed the computation with the exact NTK. Very recent preprints provide follow-on work that pushes the performance of limit kernels \citep{shankar2020without} and demonstrated the utility of limit kernels for small data tasks \citep{arora2020small}. Extending on the results for convolutional architectures, \citet{yang2019wide} showed how infinite limits could be derived for a much wider range of network architectures.

In the kernel and Gaussian process community, kernels with convolutional structure have also been proposed. Notably, these retained spatial correlation in either a fixed \citep{vdw2017convgp} or adjustable \citep{mairal2014ckn,dutordoir2020} way. While these methods were not derived using an infinite limit, \Citet{vdw2019thesis} provided an initial construction from an infinitely wide neural network limit. Inspired by these results, we propose limits of deep convolutional neural networks which retain spatial correlation in a similar way.

\section{Conclusion}
The disappearance of spatial correlations in infinitely wide limits of deep convolutional neural networks could be seen as another example of how Gaussian processes lose favourable properties of neural networks. While other work sought to remedy this problem by changing the architecture (mean-pooling), we showed that changing the weight prior could achieve the same effect. Our work has three main consequences:
\begin{enumerate}
    \item Weight correlation shows that locally connected models (without spatial correlation) and mean-pooling architectures (with spatial correlation) actually exist at ends of a spectrum. This unifies the two views in the neural network domain. We also unify two known convolutional architectures that were introduced from the Gaussian process community.
    \item We show empirically that performance improvements can be gained by using weight correlations \emph{between} the extremes of locally connected networks or mean-pooling. We also show that mean-pooling in intermediate layers can be replaced by weight correlation in infinitely wide architectures.
    \item Using weight correlation may provide advantages during hyperparameter tuning. Discrete architectural choices need to be searched through simple evaluation, while continuous parameters can use gradient-based optimisation. While we have not taken advantage of this in our current work, this may be a fruitful direction for future research.
\end{enumerate}


\begin{acknowledgements}
The authors would like to thank the reviewers for helpful comments.
AGA was supported by a UK Engineering and Physical Sciences Research Council studentship [1950008].
\end{acknowledgements}

\bibliography{cnn-limits}

\cleardoublepage

\appendix 
\section{Patch functions and discrete convolutions} \label{app:patchf}
Usually, convolutions are defined explicitly by subtracting the indices of one input tensor from the other one, and not using patch functions. To make this paper clearer, it is convenient to abstract the details of a convolution, so we introduced the patch function.

\begin{definition}[Discrete convolution]
  \label{def:convolution}
  Let $D \in \sN$ be a number of spatial dimensions, the tensor-valued weights
  $\mW\in \sR^{\patchsizebase}$, input $\mX\in \sR^{\vF}$, and output $\mY \in \sR^{\vF'}$.
  The tensor sizes $\patchsizebase$ (patch size) and $\layersizebase,\layersizebase'$ (feature sizes) are each a
  $D$-tuple, $\patchsizebase,\layersizebase,\layersizebase' \in \sN^{D}$.
  We say that $\mY$ is the result of the convolution operation $\mY = \mW*\mX$, if
  \begin{equation}
    Y_{\nextpatchs_{1},\dots,\nextpatchs_{D}} = \sum_{\patchs_{1}=1}^{P_{1}} \dots \sum_{\patchs_{D}=1}^{P_{D}} W_{\patchs_{1},\dots,\patchs_{D}} X_{\patchf{\nextpatchs_{1}}{\patchs_{1}}, \dots, \patchf{\nextpatchs_{D}}{\patchs_{D}}}.
    \label{eq:def-convolution-no-tuples}
  \end{equation}
  Here, $\patchf{\nextpatchs_{d}}{\cdot} : \countto{P_{d}} \to \countto{F'_{d}}$ are the \emph{patch functions} for a given output location $\nextpatch$. Using $D$-tuples $\patch,\nextpatch$ as indices, we may also write
  \begin{equation}
    Y_{\nextpatch} = \sum_{\patch=\1}^{\vP} W_{\patch} X_{\patchf{\nextpatch}{\patch}}.
    \label{eq:def-convolution-tuples}
  \end{equation}
  Counting from $\1$ to $\patchsizebase$ is done in such a way that $\patch$ takes all the values in $\countto{\patchsizebase}.$
  \end{definition}

  \begin{definition}[Patch function]
    \label{def:patchf}
    For each dimension $d \in \countto{D}$, layer $\ell\in\countto{L}$, fix a
    stride $s\in\sN$,
    and dilation $h \in \sN$.
   For output position $\nextpatch$, the patch function of the $d$th dimension $\patchf{\nextpatchs_{d}}{\cdot} : \countto{P_{d}} \to \countto{F'_{d}}$ is
\begin{equation}
  \patchf{\nextpatchs_{d}}{\patchs_{d}} = s\nextpatchs_{d} -  h\bra{\patchs_{d} - \ceil{\frac{\patchsizebases_{d}}{2}}} .
  \label{eq:def-patchf}
\end{equation}
For a $D$-tuple index $\patch$, we may compactly write $\patchf{\nextpatch}{\patch} \eqdef \bra{\patchf{\nextpatchs_{1}}{\patchs_{1}}, \dots, \patchf{\nextpatchs_{D}}{\patchs_{D}}}$.
\end{definition}
It is possible to verify that \cref{def:patchf} overall yields the usual definition of a convolution in deep learning \citep[Section~9.1]{deeplearningbook}.

\begin{remark}
  The concatenation of two patch functions $\patchf{\nextpatch}{\cdot},\patchf{\nextpatch'}{\cdot}$ is also a patch function, with argument in $\countto{\patchsizebase}^{2}$. That is, for $[\patch,\patch'] = \vs \in \sN^{\patchsizebase \times \patchsizebase}$ and $[\nextpatch,\nextpatch'] = \vr$,
  \begin{align}
    (&\patchf{\nextpatch}{\patch},\patchf{\nextpatch'}{\patch'}) \nonumber\\&=
    \bra{\patchf{\nextpatchs_{1}}{\patchs_{1}}, \dots, \patchf{\nextpatchs_{D}}{\patchs_{D}},
    \patchf{\nextpatchs'_{1}}{\patchs'_{1}}, \dots, \patchf{\nextpatchs'_{D}}{\patchs'_{D}}
    }\nonumber\\&= \patchf{\vr}{\vs}\,.
  \end{align}
  \label{rem:concat-patchf}
  \end{remark}

\section{Proof that a CNN with correlations in the weights converges to a GP}\label{app:netsor}

In this section, we formally prove that a CNN with correlated weights converges in distribution
to a Gaussian process in the limit of infinite width. Using the \Netsor
programming language due to \citet{yang2019wide}, most of the work in the proof
is done by
one step: describe a CNN with correlated weights in \Netsor.

For the reader's convenience, we informally recall the \Netsor programming language
\citep{yang2019wide} and key properties of its programs \crefp{thm:netsor-master,corollary:netsor-gp}.
The outline of our presentation here also closely follows
\citet{yang2019wide}. Readers familiar with \Netsor should skip to
\cref{sec:netsor-cnn-program}, where we show the program that proves \cref{thm:correlated-cnn-gp}.

\subsection{Definition of a \Netsor program}\label{sec:netsor-programming}
A \Netsor program expresses numerical computations, such as those used to define the output of a neural network. Each line of a \Netsor program is simply the definition of a new variable, in terms of previously defined variables.

There are three types of variables: $\Gva(n)$-vars, $\Ava(n_1, n_2)$-vars, and
$\Hva(n)$-vars (henceforth called ``\Netsor variables''). Each of these have one
or two parameters, which are the widths we will take to infinity. For a given
index in $\countto{n}$ (or $\countto{n_1} \times \countto{n_2}$), each \Netsor
variable is a random \emph{scalar}. To represent vectors that do not grow to
infinity, we need to use collections of \Netsor variables.

$\Gva$-vars, $\Ava$-vars and $\Hva$-vars are all random when the program is run. To accomodate non-random variables that may change (like the input $\mX$ to a \ac{NN}) we must define a different \Netsor program, defining $\mX$ either as a constant or a $\Gva$-var with variance zero.

What follows is an explanation of the three kinds of \Netsor variables, and
example uses of them. The program indicates the type of a variable using
``$\text{var} : \text{Type}$''.
\begin{itemize}
\item[$\Gva$-vars] (Gaussian-vars) are $n$-wise \emph{approximately}
\ac{iid} and Gaussian. By ``$n$-wise (approximately) independent'' we mean that there can be
correlations between $\Gva$-vars, but only within a single index $i \in 1,\dots,n$.
$\Gva$-vars will converge in
distribution to an $n$-wise independent, identically distributed Gaussian in the limit of $n \to \infty$, if
all widths are $n$. They are used, for example, to define the biases of a \ac{FCNN}.

\item[$\Ava$-vars] represent matrices,
like the weight matrices of a dense neural network. Their entries are always \ac{iid}
Gaussian with with zero mean, even for finite instantiations of the program
(finite $n$).
There are no correlations between different $\Ava$-vars, or elements of the same $\Ava$-var. They may be used to define the weight matrices of a \ac{FCNN}.

\item[$\Hva$-vars] represent variables that become $n$-wise \ac{iid} (not necessarily
Gaussian) in the
infinite limit. $\Gva$ is a subtype of $\Hva$, so all $\Gva$-vars are also $\Hva$-vars. Post-nonlinearity activations are $\Hva$-vars.

\item[$\Ova$-vars] (Output-vars) are used to define the output of the \Netsor
program. A $\Ova(n)$-var behaves like you would expect a hypothetical $\Ava(n, 1)$-var to behave: its elements are \ac{iid} Gaussian with mean zero, and it is independent of all other variables in the program.
\end{itemize}

\citet{yang2019wide} does not define $\Ova$-vars, instead choosing to consider them part of the $\Gva$-vars, since they both converge to \ac{iid} Gaussians.

\begin{definition}[Netsor program]
A \textsc{Netsor} program consists of:

\textbf{Input:}
 A set that may contain $\Gva$-vars, $\Ava$-vars, and $\Ova$-vars.

\textbf{Body:}
Each line of the program defines a new variable in terms of existing ones.
New variables may be defined using the following rules:
\begin{itemize}
  \item\MatMul: $\Ava(n_1, n_2) \times \Hva(n_2) \to \Gva(n_1)$. Given
        an $\Ava(n_{1}, n_{2})$-var (\ac{iid} Gaussian matrix) and an $\Hva(n_{2})$-var (\ac{iid} vector), its multiplication is a $\Gva(n_{1})$-var (that is, it
        converges to a Gaussian vector in the limit $n_2 \to \infty$).
  \item\LinComb: Given constants $\alpha_1,\dots,\alpha_K$, and $\Gva$-vars
    $x_1,\dots,x_K$ of type $\Gva(n_1)$, their linear combination $\sum_{k=1}^K \alpha_k
    x_k$ is a $\Gva$-var.
  \item\Nonlin: applying an elementwise nonlinear function $\phi : \sR^K \to
    \sR$, we map
    several $\Gva$-vars $x_1,\dots,x_K$ to one $\Hva$-var.
\end{itemize}

\textbf{Output:}
A tuple of scalars $(o_1^\tp x_1, \;\dots, \; o_K^\tp
x_K/\sqrt{n_K})$. The variables $o_k : \Ova(n_{k})$ are $\Ova$-vars. It may be the case that $o_j = o_k$ for
different $j, k$ (that is, the list $[v_{1},\dots,v_{K}]$ has repeated entries). Each $x_k : \Hva(n_k)$ is a $\Hva$-var.
\end{definition}

Outside of these rules, \Netsor does not have conditionals or loops\footnote{Of course, a nonlinearity $\phi$ may be internally defined using loops and conditionals, so long as it satisfies \cref{asm:controlled}.}. In
practice, we may use loops and conditionals to write a \Netsor program, so long
as they do not access the values of \Netsor variables. Conceptually, these
behave like a LISP-style ``macro'' that generates a \Netsor program.


\subsection{The output of a \Netsor program converges to a Gaussian process}\label{sec:netsor-converges}
For simplicity, we assume that the width of all the \Netsor variables is $n$.
\Citet{yang2019wide} also considers the case where each $n_{k}$ is different. First, the necessary assumptions.

\begin{definition}[Controlled function \citep{yang2019wide}]
  A function $\phi: \sR^k \to \sR$ is \emph{controlled} if it is measurable and
  \[ \abs{\phi(\vx)} \le \text{exp} \bra{C\bracket{\|}{\|}{\vx}_2^\bra{2-\epsilon}}
    + c \]
  for some
  $C,c,\epsilon > 0$, where $\|\cdot\|_2$ is the L2 norm.
  \label{def:controlled-function}
\end{definition}

If a function is controlled, it is L2 integrable with a Gaussian. That is, if
the argument $x$ of the function is Gaussian, the variance of $\phi(x)$ is
finite. This in turn ensures that the \ac{NN} function has finite variance. All
common nonlinearities (ReLU, tanh, SiLU, \dots) are controlled. This is a very
weak assumption, it is vanishingly unlikely that future nonlinearities will grow as fast
as $O\bra{e^{x^{2}}}$.

\begin{assumption}
  All nonlinear functions $\phi(\cdot)$ in the \Netsor program are controlled.
  \label{asm:controlled}
\end{assumption}

\begin{assumption}[Distribution of $\Ava$-var inputs] Consider each $\Ava(n, n)$-var in the program, $\mW$. Each of its elements
  $W_{\chan,\prevchan}$, where $\chan,\prevchan \in \countto{n}$, is  sampled from the zero-mean, \ac{iid} Gaussian, $W_{\chan,\prevchan} \simiid \Normal{0,
    \sigma_{\text{w}}^2/n}.$
  \label{asm:avar-inputs}
\end{assumption}
\begin{assumption}[Distribution of $\Gva$-var inputs]
  Consider the input vector of all $\Gva(n)$-vars for each channel $\chan \in \countto{n}$,
  that is the vector $\vz_\chan \eqdef [x_\chan : x\text{ is input
  }\Gva\text{-var}]$. It is drawn from a Gaussian, $\vz_\chan \simiid
  \Normal{\vmu^\text{in}, \vSigma^\text{in}}$.
  The covariance $\vSigma^\text{in}$ may be
  singular.\label{asm:gvar-inputs}
\end{assumption}
\begin{assumption}[Distribution of $\Ova$-vars]
  Each $\Ova(n_{k})$-var $v_{k}$ in the program is an independent Gaussian for each channel. Different $\Ova$-vars may have different variances.
  That is, for each $k\in\countto{K}, \chan\in\countto{n}$, $v_{k,\chan} \simiid \Normal{0, \sigma_{k}^{2}/n_{k}}$.
  \label{asm:ovar-inputs}
\end{assumption}

\begin{theorem}[\Netsor master theorem, \citealp{yang2019wide}]
  Fix any \Netsor program satisfying \cref{asm:controlled,asm:avar-inputs,asm:gvar-inputs,asm:ovar-inputs}. If $g\ssup{1},\dots,g\ssup{M}$ are all the $\Gva$-vars in the entire program, then for any controlled $\psi : \sR^{M} \to \sR$, as $n\to\infty$,
  \begin{multline}
    \frac{1}{n}\sum_{\chan=1}^{n} \psi\bra{g\ssup{1}_{\chan},\dots,g\ssup{M}_{\chan}}\\\asconverges \Expect{\vz \sim \Normal{\vm, \mK}}{\psi\bra{z\ssup{1}, \dots, z\ssup{M}}}.
    \label{eq:strong-lln-psi}
  \end{multline}
  Here $\asconverges$ is almost sure convergence \citep[sec.~5.2]{probability-theory-intro}. The mean $\vm$ and covariance $\mK$ are calculated under the assumption that all the $\Gva$-vars are jointly Gaussian, like in \cref{sec:deep-correlations}.
  \label{thm:netsor-master}
\end{theorem}
\begin{proof}[Proof sketch]
  The proof is by induction on the number of $\Gva$-vars included in the output,
  added in order of definition. The induction invariant is that, for some $m < M$, \cref{eq:strong-lln-psi} holds; and that a subset of $\Gva$-vars in $\countto{m}$ which form a basis have a non-singular distribution. The detailed proof is in \citet[Appendix~H]{yang2019wide}.
\end{proof}

The following corollary is a consequence of the Master theorem
(\ref{thm:netsor-master}) and the \acl{CLT}.

\begin{corollary}[Corollary~5.5, abridged, \citealp{yang2019wide}]
Fix any \Netsor program which satisfies \cref{asm:controlled,asm:avar-inputs,asm:gvar-inputs,asm:ovar-inputs}.
For simplicity, fix
  the widths of all the variables to $n$. The program outputs are $(o_{1}^\tp x_1, \;\dots, \; v_K^\tp
  x_K)$, where
  each $x_k$ is an $\Hva$-var, and each
    $o_k$ is a $\Ova$-var.
  Then, as $n \to \infty$, the output tuple
  converges in distribution to a Gaussian $\Normal{{\boldsymbol{0}}, \vK}$.
  The covariance $\vK$ is given by doing calculations like \cref{sec:deep-correlations}, assuming that $\Gva$-vars are jointly Gaussian.
  \label{corollary:netsor-gp}
\end{corollary}

\subsection{\Netsor program and GP behaviour: CNN with correlated weights}\label{sec:netsor-cnn-program}
\Netsor only has native support for matrix-vector multiplications and linear
combinations with constants. How can we represent a convolution operation for \ac{CNN}? Consider the convolutional layer definition \eqparref{eq:deep-recursion}. Changing the sum order, we obtain
\begin{align}
  \intertext{Expanding the convolution into a sum, and changing the sum order, we obtain}
 \layerAsm{\ell}{\chan,\nextpatch}{\mX} &= \sum_{\patch=\1}^{\patchsize{\ell}} \sum_{\prevchan=1}^{\layerC{\ell-1}}\layerWs{\ell}_{\chan,\prevchan,\patch}
                         \;{\layernlasm{\ell-1}{\prevchan,\patchf{\nextpatch}{\patch}}{\mX}},
\intertext{which is just a spatial sum of matrix multiplications}
\layera{\ell}{:,\nextpatch} &= \sum_{\patch=\1}^{\patchsize{\ell}} \layerW{\ell}_{:,:,\patch} \; \layernla{\ell-1}{:,\patchf{\nextpatch}{\patch}}.
                              \label{eq:conv-as-matmul}
\end{align}
Thus, we may express a convolution with multiple filters as a sum of matrix-vector multiplications. This is
the canonical way to represent
convolutional filters in \Netsor \citep[\Netsor program~4]{yang2019wide}.

Here we run into a problem. \Cref{eq:correlated-weight-prior} states that \ac{CNN}
filters are spatially correlated, but \cref{asm:avar-inputs} states that
$\Ava$-vars have to be independent. To solve this, we will use the the following
well-known lemma, which is the $\mR\epsilon$ expression of a Gaussian random
variable with mean zero. The tensor $\mR$ is a square root of the covariance.

\begin{lemma}
  Let $\mSigma \in \sR^{\patchsizebase^{2}}$ be an arbitrary real-valued covariance tensor.
  Then there exists another real-valued tensor $\mR \in \sR^{\patchsizebase^{2}}$ such that $\Sigma_{\nextpatch,\nextpatch'} = \sum_{\patch=1}^{\patchsizebase}R_{\nextpatch,\patch} R_{\nextpatch',\patch}$.
  Next, let $\vu,\vw \in \sR^{\patchsizebase}$ be real-valued tensors, such that $\vw = \mR \vu$. Suppose the elements of $\vu$ are \ac{iid} standard Gaussian variables, $\cb{u_\patch}_{\patch \in \countto{\patchsizebase}} \simiid \Normal{0, 1}$. Then, $\vw$ has a multivariate Gaussian distribution with mean zero and covariance tensor $\vSigma$.
  \label{lemma:R}
\end{lemma}
\begin{proof}
  Let $K = \abs{\patchsizebase}$, and $\tilde\mSigma$ be $K \times K$ matrices, obtained by flattening the dimensions of $\mSigma$ respectively.
  Then $\tilde\mSigma$ is a real-valued covariance matrix, so it is positive semi-definite and
  a square matrix $\tilde\mR$ s.t. $\tilde\mR{\tilde\mR}^\tp = \tilde\mSigma$ always exists. Un-flattening $\tilde\mR$ we obtain $\mR$. The variable $\vw$ is Gaussian because it is
  a linear transformation of the Gaussian $\vu$. Calculating the second moment of $\vw$ finishes the proof.
\end{proof}
Thus, to express convolution in \Netsor with correlated weights $\vw$, we can use the
following strategy. First, express several convolutions with uncorrelated
weights $\vu$, using \cref{eq:conv-as-matmul}. Then, combine the output of the convolutions using \LinComb and coefficients of
the tensor $\mR$.

Given a collection of $\Ava$-vars
$\cb{\layerU{\ell}_{:,:,\patch}}_{\patch \in \countto{\patchsize{\ell}}}$, we
can express the convolutional weights $\layerW{\ell}$ which have covariance
$\priorWcov{\ell} = \mR\ssup{\ell} \bra{\mR\ssup{\ell}}^{\tp}$ as
\begin{equation}
  \layerW{\ell}_{:,:,\patch} = \sum_{\vs=\1}^{\patchsize{\ell}}
  R\ssup{\ell}_{\patch,\vs}\layerU{\ell}_{:,:,\vs}\,.
\end{equation}

Substituting this into \cref{eq:conv-as-matmul} we obtain
\begin{equation}
\layera{\ell}{:,\nextpatch} = \sum_{\patch=\1}^{\patchsize{\ell}} \sum_{\vs=\1}^{\patchsize{\ell}}R\ssup{\ell}_{\patch,\vs}\layerU{\ell}_{:,:,\vs} \; \layernla{\ell-1}{:,\patchf{\nextpatch}{\patch}}.
\label{eq:correlated-conv-as-matmul}
\end{equation}
To express this computation with \Netsor rules we may write
\begin{align}
  \texttt{MatMul: }& \mH\ssup{\ell}_{:,\vs,\patch}\bra{\mX} \eqdef \layerU{\ell}_{:,:,\vs}\layernla{\ell-1}{:,\patch} \nonumber\\
  &\text{for }\vs\in \countto{\patchsize{\ell}},\patch \in \countto{\layersize{\ell-1}}, \label{eq:cnn-corr-matmul} \\
  \texttt{LinComb: }& \layera{\ell}{:,\nextpatch} \eqdef +
\sum_{\patch=\1}^{\patchsize{\ell}} \sum_{\vs=\1}^{\patchsize{\ell}}
R\ssup{\ell}_{\patch,\vs}\;
\mH\ssup{\ell}_{:,\vs,\patchf{\nextpatch}{\patch}}\bra{\mX}  \nonumber\\
  &\text{for }\nextpatch\in \countto{\layersize{\ell}}.
\label{eq:cnn-corr-lincomb}
\end{align}
\Cref{alg:netsor-cnn} uses this construction for every layer to express an
$L$-layer \ac{CNN} with correlated weights, applied to an input data set
$\gX\eqdef\sqb{\mX_{1},\dots,\mX_{M}}$.

\begin{algorithm}[tp]
\SetKwInOut{Input}{Input}
\SetKwInOut{Output}{Output}
  \caption{\Netsor description of an $L$-layer CNN with
    correlated weights, with input $\gX$.}\label{alg:netsor-cnn}
  \CommentC{$\Gva$-vars for layer 1 activations, for all spatial locations
     $\patch$ and input points $\mX_m$.}
  \SInput{$\layerAsm{1}{\patch}{\mX_m} : \Gva(\layerC{1})$ \newline for $\patch \in
  \countto{\layersize{1}}$ and $m =1,\dots,M.$}
  \CommentC{$\Ava$-vars for the independent convolutional weights}
  \SInput{$\layerU{\ell}_{\patch} : \Ava\bra{\layerC{\ell}, \layerC{\ell-1}}$\newline
      for $\patch \in [\patchsize{\ell}]$ and $\ell=2,\dots,L-1$.}
  \CommentC{$\Ova$-vars for the output, for every patch location $\vs$ and channel $\chan$}
  \SInput{$\vo_{\chan,\vs} : \Ova\bra{\layerC{L-1}}$ \newline for $\vs \in \countto{\patchsize{L}}$ and $\chan =1,\dots,\layerC{L}.$}
  \vspace{1ex}

  \For{$m = 1, \dots, M$ (data points of $m$)}{
  \For{$\ell = 2,\dots,L-1$ (layer $\ell$)}{
  \For{$\patch = \1,\dots,\layersize{\ell-1}$}{
  \SNonlin{$\Hva\bra{\layerC{\ell-1}}$ \newline $\layernlasm{\ell-1}{:,\patch}{\mX_{m}} \eqdef \phi\bra{\layeram{\ell-1}{:,\patch}{\mX_{m}}}$}

  \For{$\vs = \1,\dots,\patchsize{\ell}$ (patch location $\vs$)}{
  \SMatMul{$\Gva\bra{\layerC{\ell}}$ \newline $\mH\ssup{\ell}_{:,\vs,\patch}\bra{\mX_{m}} \eqdef \layerU{\ell}_{\vs}\layernlasm{\ell-1}{\patch}{\mX_{m}}$}
  } }
  \For{$\nextpatch=\1,\dots,\layersize{\ell}$ (spatial location $\nextpatch$)}{
  \SLinComb{$ \Gva\bra{\layerC{\ell}}: \layeram{\ell}{:,\nextpatch}{\mX_{m}}$ \newline $\eqdef
\sum_{\patch=\1}^{\patchsize{\ell}} \sum_{\vs=\1}^{\patchsize{\ell}}
R\ssup{\ell}_{\patch,\vs}\;
\mH\ssup{\ell}_{:,\vs,\patchf{\nextpatch}{\patch}}\bra{\mX_{m}} $}
  } }
  \vspace{1ex}
  \For{$\patch \in \sqb{\layersize{L-1}}$ (spatial location $\patch$)}{
  \SNonlin{$ \Hva\bra{\layerC{L-1}}$ \newline $\layernlasm{L-1}{:,\patch}{\mX_{m}} \eqdef \phi\bra{\layeram{L-1}{:,\patch}{\mX_{m}}}$}
  }
  }

  \CommentC{One output for every spatial location $\patch$, patch location $\vs$, channel $i$ and data point $m$.}
  \SOutput{$\big(\vo_{\chan,\vs}^\tp \layernlasm{L-1}{:,\patch}{\vX_m} : \text{ for
    }\patch \in \countto{\layersize{L}}, \vs\in\countto{\patchsize{L}},$\newline
    $\chan \in \countto{\layerC{L}}\text{ and }m \in \countto{M}\big)$}
  \vspace{1ex}
  {\textbf{Output postprocessing: } correlate the outputs and add biases (not part of \Netsor)}

  \For{$m \in \countto{M}$, $\chan \in \countto{\layerC{L}}$ and $\nextpatch \in \layersize{L}$}{
  {$\layerAsm{L}{\chan,\nextpatch}{\vX_m} \eqdef
      \sum_{\patch=\1}^{\patchsize{L}} \sum_{\vs=\1}^{\patchsize{L}} R\ssup{L}_{\patch,\vs} \;
      \bra{\vo_{\chan,\vs}^\tp \layernlasm{L-1}{:,\patchf{\nextpatch}{\patch}}{\vX_m} }$}
    }
\end{algorithm}

\begin{theorem}[Correlated CNN behaves like a GP]
  Consider a countable set of input points $\tilde{\gX}$, and a fixed number of
  layers $L$. Apply the $L$-layer convolutional neural network \crefp{eq:deep-recursion,eq:deep-activations}
  with correlated weights \crefp{eq:correlated-weight-prior} to $\gX$. Assume its
  nonlinearities are controlled \crefp{asm:controlled}.
  For simplicity, fix all layers to have
  the same number of channels: $C = \layerC{1} = \dots = \layerC{L}$. Then, as the number of channels $C \to \infty$, the activations $\layerAm{L}{}{\tilde\gX}$ converge in distribution to a Gaussian process with mean function $\E\sqb{\layerAsm{L}{\chan}{\tilde\gX}} = \meanf{L}\bra{\tilde{\gX}}$ and covariance function $\Cov\sqb{\layerAsm{L}{\chan,\nextpatch}{\tilde\gX}, \layerAsm{L}{\chan',\nextpatch'}{\tilde{\gX}}} =
  \delta_{\chan,\chan'}\covf{L}_{\nextpatch,\nextpatch'}\bra{\tilde\gX,\tilde\gX}$ \crefp{sec:deep-correlations}.
\label{thm:correlated-cnn-gp}
\end{theorem}
\begin{proof}
  We need to show
  \begin{enumerate}
    \item that \cref{alg:netsor-cnn}, including the postprocessing part, implements a correlated-weight CNN correctly,
    \item that the full program converges weakly to a \ac{GP} on $\tilde\gX$,
    \item that the moments of this \ac{GP} match the ones in \cref{sec:deep-correlations}.
  \end{enumerate}

  For the first claim, the key is the equivalence between a convolutional layer
  with correlated weights \eqparref{eq:deep-recursion}, and a spatial outer
  product followed by linear combination
  \eqparreftwo{eq:cnn-corr-matmul}{eq:cnn-corr-lincomb}. Keeping this in mind, we
  can verify by inspection that the steps of \cref{alg:netsor-cnn}, including
  the output postprocessing, implement the recursive CNN equations
  \crefp{eq:deep-recursion,eq:deep-activations}.

  The second claim is somewhat more involved. Invoking the Kolmogorov extension
  theorem \citep[Thm.~2.4.3]{measure-theory-intro} we restrict our attention to finite
  subsets $\gX \subseteq \tilde\gX$, which are going to be compatible
  distributions if claim~3 is true. Since $\mX \in \gX$ are tensors, we may use the Euclidean metric.
  We then show by \cref{thm:netsor-master} that the output tuple of the CNN
  \Netsor program converges weakly to a GP as $C\to\infty$. Each activation in
  the postprocessing is defined as a linear combination of a Gaussian \ac{RV}
  (the bias) and a \ac{RV} that converges weakly to a Gaussian, and thus the resulting distribution on $\gX$ converges weakly to a Gaussian too.

  Finally, we have to show that the postprocessed output has the correct kernel.
  The output tuple and the activations have mean zero, which is correct. We thus compute the covariance of the output tuple in \cref{alg:netsor-cnn}, for $\mX,\mX' \in \tilde\gX$:
  \begin{multline}
    \Cov\sqb{
      \vo_{\chan,\vs}^\tp \layernlasm{L-1}{:,\patch}{\mX},
      \vo_{\chan',\vs'}^\tp \layernlasm{L-1}{:,\patch'}{\mX'}
    } \\= \delta_{\chan,\chan'} \delta_{\vs,\vs'} \nlinf{L-1}_{\patch,\patch'}\bra{\mX,\mX'}.
  \end{multline}
  The delta functions appear because the $\Ova$-vars and their elements are all independent. Using this, we can calculate the covariance function of the activations
  \begin{multline}
\Cov\sqb{
  \layerAsm{L}{\chan,\nextpatch}{\vX}, \layerAsm{L}{\chan',\nextpatch'}{\vX'}}
= \sum_{\patch,\patch'}^{\patchsize{L}^{2}}\sum_{\vs,\vs'}^{\patchsize{L}^{2}}\\
R\ssup{L}_{\patch,\vs} R\ssup{L}_{\patch',\vs'}
\Cov\sqb{
      \vo_{\chan,\vs}^\tp \layernlasm{L-1}{:,\patchf{\nextpatch}{\patch}}{\mX},
      \vo_{\chan',\vs'}^\tp \layernlasm{L-1}{:,\patchf{\nextpatch'}{\patch'}}{\mX'}
    }.
\end{multline}
Substitute the value of the expectations and eliminate one of the sums due to $\delta_{\vs,\vs'}$,
\begin{align}
&\Cov\sqb{
  \layerAsm{L}{\chan,\nextpatch}{\vX}, \layerAsm{L}{\chan',\nextpatch'}{\vX'}} \nonumber\\
&=\delta_{\chan,\chan'} \big(\sum_{\patch,\patch'}^{\patchsize{L}^{2}}\sum_{\vs=\1}^{\patchsize{L}}
   R\ssup{L}_{\patch,\vs} R\ssup{L}_{\patch',\vs} \nlinf{L-1}_{\patchf{\nextpatch}{\patch},\patchf{\nextpatch'}{\patch'}}\bra{\mX,\mX'} \big).
\intertext{
Finally, noting that $\priorWcov{L}_{\patch,\patch'} = \sum_{\vs=\1}^{\patchsize{L}} R\ssup{L}_{\patch,\vs} R\ssup{L}_{\patch',\vs} $ \crefp{lemma:R}, and using the definition of $\covf{L}_{\nextpatch,\nextpatch'}\bra{\mX,\mX'}$ \eqparref{eq:cnn-kernel-recursive}, we obtain the claim.
}
  &= \delta_{\chan,\chan'} \big(
\sum_{\patch,\patch'}^{\patchsize{L}^{2}} \priorWcov{L}_{\patch,\patch'} \nlinf{L-1}_{\patchf{\nextpatch}{\patch},\patchf{\nextpatch'}{\patch'}}\bra{\mX,\mX'}
\big)  \nonumber\\
&= \delta_{\chan,\chan'} \covf{L}_{\nextpatch,\nextpatch'}\bra{\mX,\mX'}.
\end{align}
\end{proof}

\end{document}